\documentclass[a4paper,fleqn]{cas-dc}

\usepackage[authoryear]{natbib}
\usepackage[strings]{underscore}
\usepackage{hyperref}
\hypersetup
{
colorlinks=true,
linkcolor=blue,
filecolor=blue,
urlcolor=blue,
citecolor=cyan,
}
\usepackage{multirow} 
\usepackage{longtable} 
\usepackage{array} 
\usepackage{makecell} 
\usepackage{threeparttable}
\usepackage{booktabs}
\usepackage{comment}
\usepackage{tablefootnote}
\usepackage[utf8]{inputenc}
\usepackage[T1]{fontenc}
\usepackage{booktabs}
\usepackage[justification=centering]{caption}
\usepackage{array, caption, threeparttable}
\usepackage[font=small,labelfont=bf,labelsep=none]{caption}
\def\tsc#1{\csdef{#1}{\textsc{\lowercase{#1}}\xspace}}
\tsc{WGM}
\tsc{QE}
\begin{document}
\let\WriteBookmarks\relax
\def\floatpagepagefraction{1}
\def\textpagefraction{.001}

\shorttitle{Densely connected neural networks for nonlinear regression}    

\shortauthors{C.Jiang et al.}

\title [mode = title]{Densely connected neural networks for nonlinear regression}  

% \tnotemark[] 

% \tnotetext[]{} 


\author[2]{Chao Jiang}

% \ead{chaojiang0728@gmail.com}


\credit{Methodology, Manuscript preparation, Manuscript submission, Formal analysis, Application design, Visualization}


\author[3]{Canchen Jiang}



% \ead{Canchen.jiang@monash.edu}

% \ead[url]{<URL>}

\credit{Methodology, Manuscript preparation, Manuscript submission, Formal analysis, Application design, Visualization}

\author[1]{Dongwei Chen}[orcid=0000-0003-1956-7683]

% Footnote of the second author
\cormark[1]
% Email id of the second author
% \ead{dongwec@g.clemson.edu}
\credit{Conceptualization, Methodology, Algorithm implementation, Manuscript preparation, Manuscript submission, Formal analysis, Application design, Data acquisition, Visualization}

\author[4,5]{Fei Hu}



% \ead{hufei@mail.iap.ac.cn}

% \ead[url]{<URL>}

\credit{Methodology, Manuscript preparation, Manuscript submission, Application design, Data acquisition}


\affiliation[1]{organization={School of Mathematical and Statistical Sciences, Clemson University},
            %addressline={}, 
            city={Clemson},
%          citysep={}, % Uncomment if no comma needed between city and postcode
            postcode={29641}, 
            state={SC},
            country={USA}}
            
\affiliation[2]{organization={Department of Civil Engineering, Faculty of Engineering, Monash University},
            %addressline={}, 
            city={Melbourne},
%          citysep={}, % Uncomment if no comma needed between city and postcode
            postcode={3800}, 
            state={VIC},
            country={Australia}}
            
\affiliation[3]{organization={Department of Data Science and Artificial Intelligence, Faculty of Information Technology, Monash University},
            %addressline={}, 
            city={Melbourne},
%          citysep={}, % Uncomment if no comma needed between city and postcode
            postcode={3800}, 
            state={VIC},
            country={Australia}}
            
\affiliation[4]{organization={Institute of Atmospheric Physics, Chinese Academy of Sciences},
            %addressline={}, 
            city={Beijing},
%          citysep={}, % Uncomment if no comma needed between city and postcode
            postcode={100029}, 
            %state={},
            country={China}}

\affiliation[5]{organization={College of Earth and Planetary Sciences, University of Chinese Academy of Sciences},
            %addressline={}, 
            city={Beijing},
%          citysep={}, % Uncomment if no comma needed between city and postcode
            postcode={100049}, 
            %state={},
            country={China}}

\cortext[1]{Corresponding author \\\indent \indent Email address: dongwec@g.clemson.edu(Dongwei Chen)}

% \fntext[1]{Corresponding author.}

% \nonumnote{dongwec@g.clemson.edu}

\begin{abstract}
Densely connected convolutional networks (DenseNet) behave well in image processing. However, for regression tasks, convolutional DenseNet may lose essential information from independent input features. To tackle this issue, we propose a novel DenseNet regression model where convolution and pooling layers are replaced by fully connected layers and the original concatenation shortcuts are maintained to reuse the feature. To investigate the effects of depth and input dimension of proposed model, careful validations are performed by extensive numerical simulation. The results give an optimal depth (19) and recommend a limited input dimension (under 200). Furthermore, compared with the baseline models including support vector regression, decision tree regression, and residual regression, our proposed model with the optimal depth performs best. Ultimately, DenseNet regression is applied to predict relative humidity, and the outcome shows a high correlation (0.91) with observations, which indicates that our model could advance environmental data analysis.
\end{abstract}


\begin{highlights}
\item Based on DenseNet, the novel regression model replaces convolution and pooling layers with fully connected layers in the dense block, and the building blocks of DenseNet regression are densely connected by concatenation shortcuts.  

\item The impacts of depth and input dimension are evaluated by numerical validation on simulated data. The results demonstrate an optimal depth (19) of neural network and recommend a limited input dimension (under 200) for DenseNet regression.

\item DenseNet regression outperforms the baseline models on simulated data, which verifies that dense concatenation shortcuts realize feature reuse and play an essential role in DenseNet regression. 

\item DenseNet regression is employed to predict relative humidity, and the outcome is highly correlated with observations ($\rho \approx 0.91$), which indicates that our model could advance environmental science.
\end{highlights}

% Keywords
\begin{keywords}
 \sep Neural networks 
 \sep DenseNet 
 \sep Concatenation shortcuts
 \sep Feature reuse
 \sep Nonlinear regression
 \sep Relative humidity prediction
\end{keywords}

\maketitle

% Main text
\section{Introduction}\label{1}
With the increasing trend in environmental datasets size and complexity, data science has become popular in environmental applications \citep{j9gibert2018environmental}. In environmental data analysis, regression is a useful technique in prediction. Many studies focus on forecasting environmental parameters to address environmental issues, incorporating air pollution, climate change, and global warming. For example, \cite{j10rosenlund2008comparison} predict nitrogen dioxide concentrations with a land use regression method to obtain the spatial distribution of traffic-related air pollution in Rome. \cite{j11rehana2019river} forecasts the river water temperature by regression to further analyze the possibility of future projections considering climate change. \cite{j12krishna2020multiparametric} propose a regression approach for greenhouse gases estimation.

Regression analysis statistically models the relationship between a dependent variable and independent variables. Linear model is the most common form of regression analysis. It is used to model linear relationships and includes general linear regression \citep{1pandey2020multiple}, stepwise linear regression \citep{2zhou2011stepwise}, linear regression with penalties like ridge regression \citep{3ahn2012using}, lasso regression \citep{4tibshirani1996regression} and elastic net regression \citep{5zou2005regularization}. However, nonlinear relationships are more common and complicated in the real world. Therefore, nonlinear regression analysis gains a lot of attention \citep{8yagiz2010application,7majda2012physics,6rhinehart2016nonlinear}. There exist many tools to model nonlinear relationships, such as Polynomial Regression \citep{9ostertagova2012modelling}, Support Vector Regression (SVR) \citep{10smola2004tutorial}, and Decision Tree Regression (DTR) \citep{11loh2011classification}, where SVR and DTR are popular nonlinear regression techniques. Nevertheless, SVR often takes a long time to be trained on large datasets, and DTR is 
extremely non-robust and NP-complete to learn an optimal decision tree \citep{dd7chen2020deep}. From the late 1980s, people began to use artificial neural networks (ANNs) for nonlinear regression since a neural network with a single hidden layer can approximate any continuous function with compact support to arbitrary accuracy when the width goes to infinity \citep{13hornik1989multilayer,14funahashi1989approximate,15kuurkova1992kolmogorov}. According to the universal approximation theorem, the regression accuracy of ANNs heavily depends on the width of the single hidden layer. However, the impact of depth on the accuracy of neural networks are not considered in this classical theorem \citep{j5chui1996limitations}.

Different from ANNs with a single hidden layer, deep neural networks (DNNs) trend in increasing the number of depth (layers) of neural networks, aiming at significantly improving the accuracy of models. In the last decade, there emerge many works on regression tasks using the deep learning-based method. Specifically, \cite{j6xu2014regression} propose a nonlinear regression approach based on DNNs to mimic the function between noisy and clean speech signals to improve speech. \cite{j7khaki2019crop} design a DNN model to accurately predict the crop yield, and the result shows that the regression behavior of DNN-based model in this scenario is better than shallow neural networks (SNNs). \cite{dd6lathuiliere2019comprehensive} conduct a comprehensive analysis of vanilla deep regression considering a large number of deep models have no significant difference in the network architecture and data pre-processing. \cite{dd7chen2020deep} develop a nonlinear regression model with the technique of ResNet. By comparing to other nonlinear regression techniques, this work indicates the nonlinear regression model based on DNNs is stable and appliable in practice. Although DNNs show the progresses in regression with the deep hidden layers, it is limited that the feature of each layers only use one time and the feature reuse is not considered to improve the nonlinear approximation capacity \citep{jj1sun2018resinnet}.

Densely connected convolutional networks (DenseNet) introduce the concatenation shortcut into its network \citep{18huang2017densely}. The concatenation shortcuts plays an significant role in realizing the feature reuse and the key information of initial input could be reserved and transmit to the output, which makes DenseNet achieve good performance in applications \citep{jj2zhang2018sparse,jj3saleh2019real,jj4zhang2020high}. DenseNet performs well in image processing because convolution is suitable for feature extraction of images with multiple channels. Usually, the input of images is high dimensional, complex, and includes redundant information \citep{danasingh2020identifying}. For instance, the input of a 96×96 pixels image with three channels would have 27,648 dimensions. If the number of neurons of the first hidden layer in the fully connected layer is the same as the input dimensions, the number of weights would be close to $10^8$, which is too enormous and aggravates the computation efficiency significantly. Convolution kernel is designed for reducing the repetitive parts among the variables and extract featured information. Hence, for image processing, convolution plays a significant role in reducing redundant information and improving the efficiency of the algorithm. However, \cite{dd7chen2020deep} find that convolutional neural networks may lose essential information from input features due to local convolution kernels and thus are not suitable for nonlinear regression. To tackle this issue, they introduce the so-called residual regression model by replacing convolution and pooling layers into fully connected (FC) layers in ResNet. By maintaining the shortcut within residual blocks, residual regression enhances data flow in the neural network and has been applied in many fields, such as computational fluid dynamics \citep{dd1rojek2021ai,dd2shin2021data}, computer-aided geometric design \citep{dd3scholz2021parameterization}, and safety control in visual servoing applications \citep{dd4shi2020deep,dd5shi2021bayesian}. Besides the localness of convolution, the independence of input features also requires the replacement of convolution layers in a neural network regression model. To sum up, the motivations of replacing convolution layers into FC layers in this work are explained as follows

\begin{figure}[ht]
\centering
\includegraphics[width=8cm,height=9cm,scale=1]{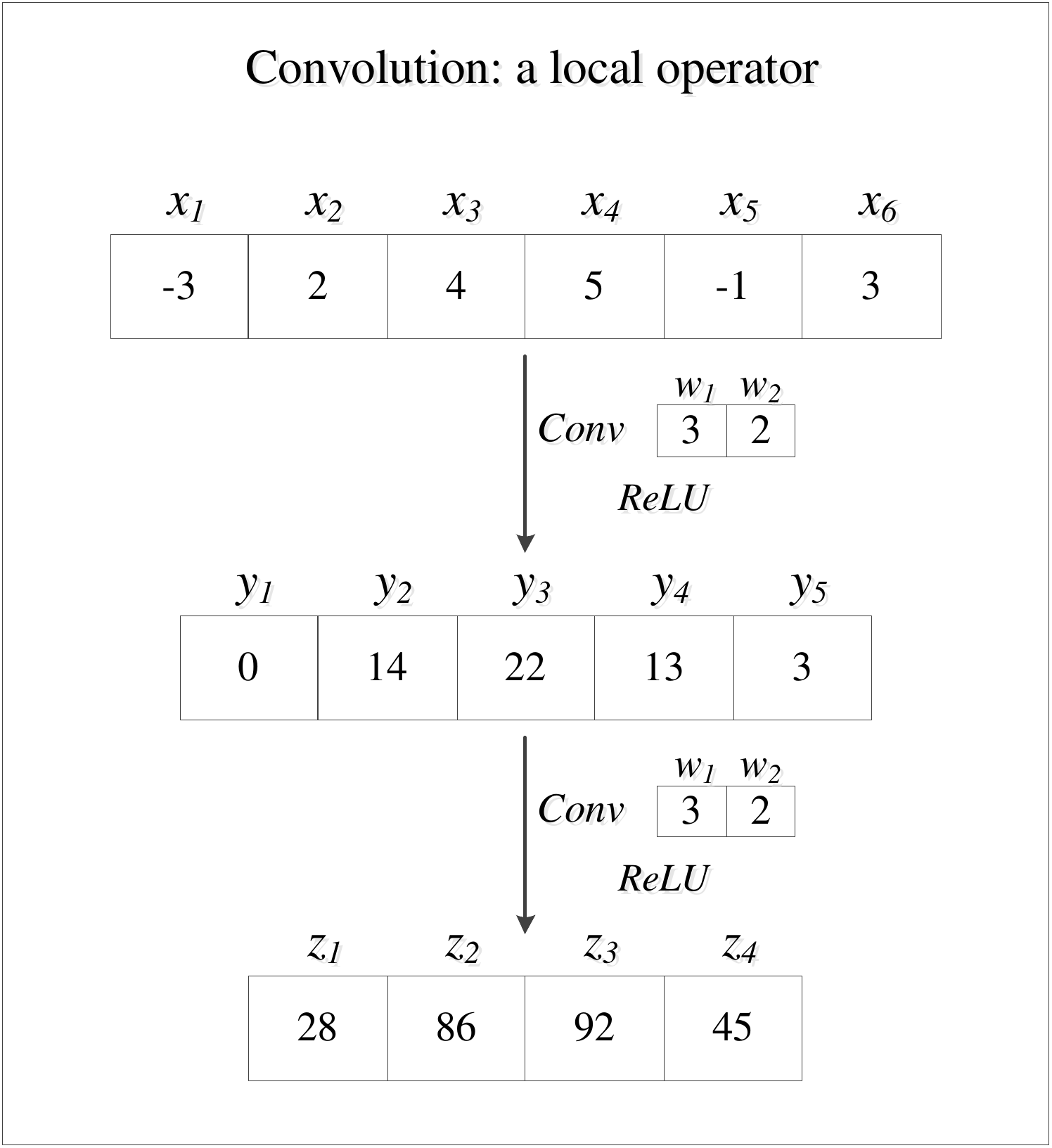}
\caption{Convolution as a local operator. The input data are $\left( x_1, \dots, x_6 \right) $ and are convoluted with a convolution kernel $\left( w_1, w_2 \right)$. The stride of convolution is 1. Then $ y_1=\text{ReLU}\left( w_1x_1 + w_2x_2 \right)=0 $. When the vector $\left( y_1, \dots, y_6 \right) $ is convoluted again, we get $ z_1=\text{ReLU}\left( w_1y_1 + w_2y_2 \right)=\text{ReLU}\left( w_2y_2 \right) $. Here when computing $z_1$, the neural network is losing information from $x_1$ since $y_1=0$. This illustrates that the localness of convolution may result in the neural network missing some input variables or essential information from input features. Thus the convolutional neural network is not suitable for regression tasks. }
\label{fig:label1}
\end{figure}

\begin{itemize}
 \item Convolution is a local operator. As introduced by \cite{dd7chen2020deep} and presented in \autoref{fig:label1}, convolution is a local operator. The localness of convolution may result in the convolutional neural network losing essential information and even key input features from input variables. Therefore, the convolutonal network is not good enough for nonlinear regression. 
 \item Predictors of regression tasks are independent. However, convolution kernels are used to extract features from redundant information or correlated variables. Therefore, the convolutional neural network would lose key information from regression predictors. Specifically, the convolutional neural network often uses 2D convolution layers. The input one-dimensional feature vector needs to be reshaped into a matrix when the network takes a regression task. The entries in the reshaped matrix are seen as the corresponding gray values of pixels of a figure and all the input entries are independent. However, the neighboring gray values in a figure are often highly correlated. This dilemma requires the substitution of convolution layers into FC layers. 
\end{itemize} 

Inspired by residual regression model, we propose a novel DenseNet model for nonlinear regression. Specifically, the new neural network retains the major architecture of DenseNet excluding convolution and pooling layers. Fully connected layers are the substitution of convolution and pooling layers in the dense block. Therefore, the conceptual architecture of our DenseNet regression model consists of a number of building blocks, and each building block is linked to the others by concatenation shortcuts. Through concatenation, the DenseNet regression model could realize feature reuse, and critical information could be reserved.

This paper is organized as follows. In \autoref{section2}, we clarify the architecture of DenseNet regression and in \autoref{section3}, we introduce the simulated dataset. In \autoref{section4}, we derive the result and have a discussion. Firstly, we evaluate the performance of DenseNet with different depths, and then we compare the results of optimal DenseNet regression model with other regression techniques. At the end of this section, we estimate the effect of input dimension on the performance of DenseNet regression. In \autoref{section5}, we use DenseNet regression to predict relative humidity. Finally, we conclude and propose the future work in \autoref{section6}.

\section{The architecture of DenseNet regression} \label{section2}
%\section{Methodology: the architecture of DenseNet regression model}
%\subsection{The architecture of DenseNet regression model}
%DenseNet is a type of high-level neural networks, aiming to improve the vanishing gradient issue in deep networks. \cite{j1shaik2021automatic} point out that vanishing gradient may especially occur in deep networks as the distance between input and output layer is longer than other networks, and information is prone to vanishing before reaching the output layer. Accordingly, based on ResNet, 

DenseNet introduces the concatenation shortcuts to enhance the feature reuse in each dense block, which is beneficial to reduce the possibility of losing critical information and increase the accuracy of DNNs.

%DenseNet performs well in image processing because convolution is suitable for feature extraction of images with multiple channels. Usually, the input of images is high dimensional, complex, and includes redundant information \citep{danasingh2020identifying}. For instance, the input of a 96×96 pixels image with three channels would have 27,648 dimensions. If the number of neurons of the first hidden layer in the fully connected layer is the same as the input dimensions, the number of weights would be close to $10^8$, which is too enormous and aggravates the computation efficiency significantly. Convolution kernel is designed for reducing the repetitive parts among the variables and extract featured information. \textcolor{blue}{Hence}, for image processing, convolution plays a significant role in improving the efficiency of the algorithm. However, for regression tasks, the input variables are independent and low-dimensional to a significant extension. There is a high probability of losing significant features if applying convolution to the independent and lower-dimensional input, as demonstrated in \autoref{fig:label1}. Therefore, the primary structure of DenseNet is not adaptive for nonlinear regression.

\begin{figure}[ht]
\centering
\includegraphics[width=5cm,height=8cm,scale=1]{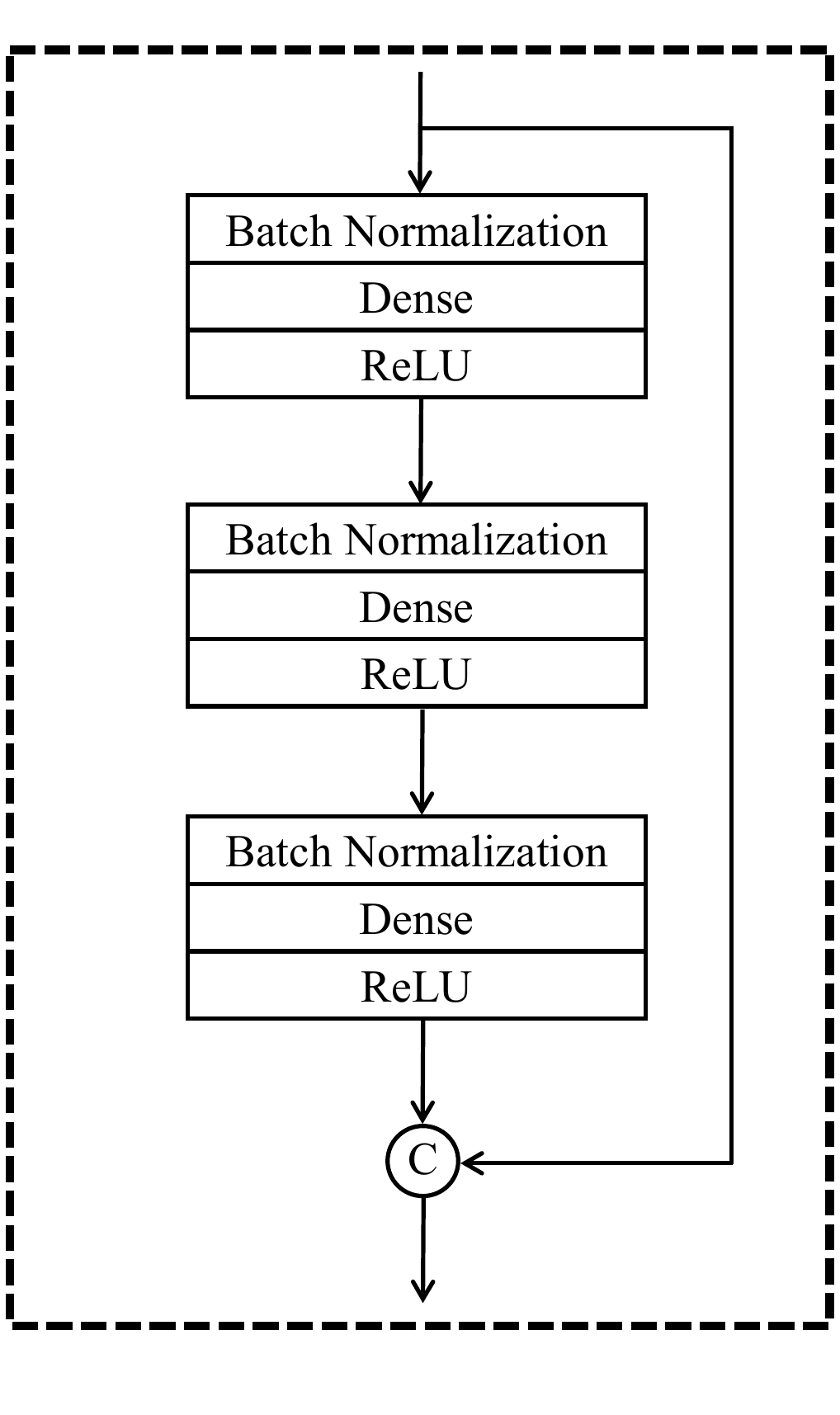}
\caption{The building block of DenseNet for nonlinear regression. There are three hidden layers in each building block. At the end of building block, there is a concatenation shortcut transmitting the information from top to bottom to guarantee feature reuse. ‘C’ in the diagram stands for concatenation.}
\label{fig:label2}
\end{figure}

\begin{figure}[ht]
\centering
\includegraphics[width=6cm,height=8cm,scale=1]{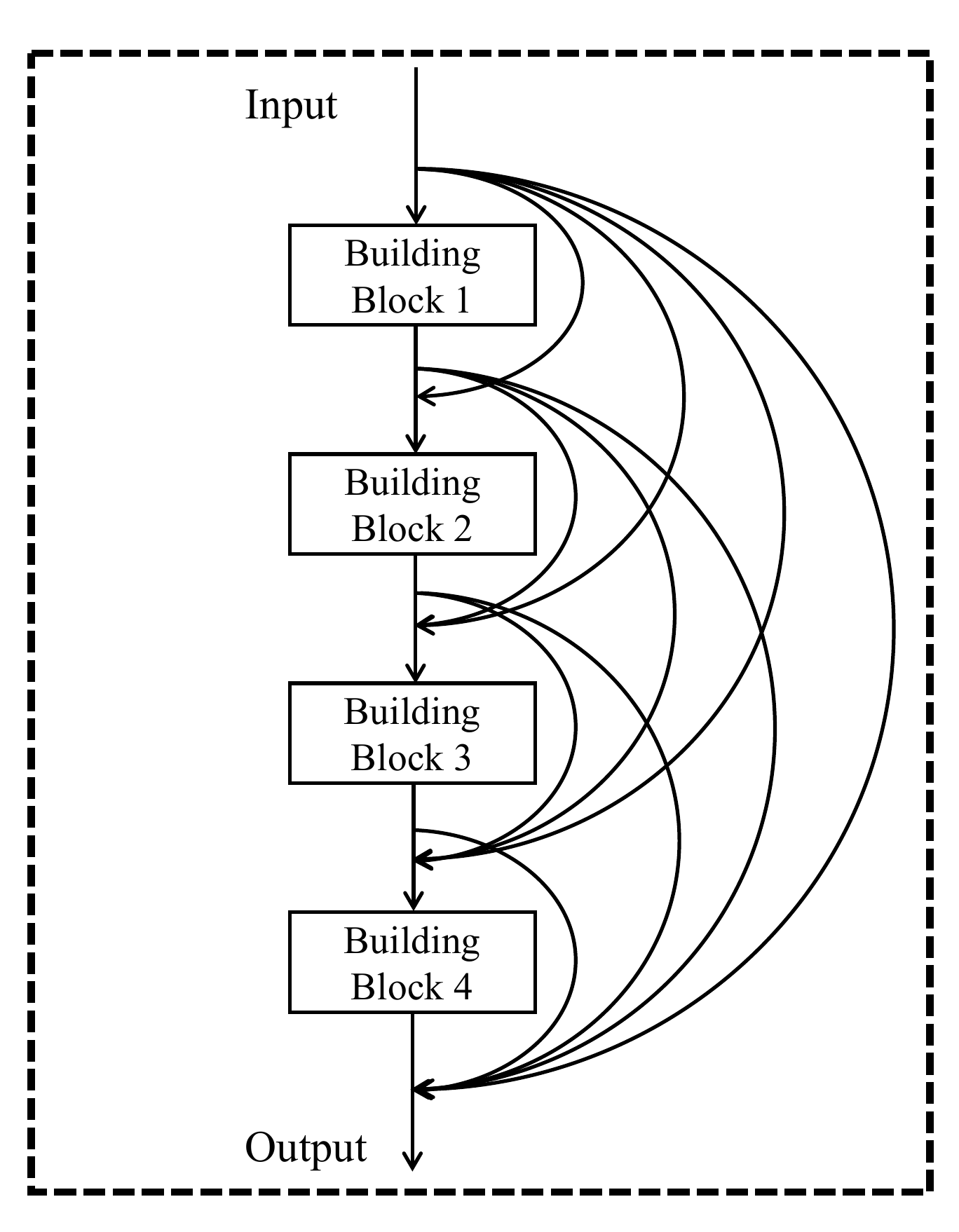}
\caption{The architecture of a DenseNet regression model with 13 layers. The model has four building blocks. Each building block is linked to the others by a concatenation shortcut. The output layer is positioned in the end. Through concatenation, the features of initial inputs and outputs of each building block would be transmitted to the following output layer.}
\label{fig:label3}
\end{figure}

Generally, based on the characteristics of DenseNet, the architecture of our proposed model employs concatenation and removes convolutional part, so that DenseNet could better serve the nonlinear regression tasks. \autoref{fig:label2} and \autoref{fig:label3} demonstrates the details of architecture of this DenseNet regression model. Unlike convolutional DenseNet, the DenseNet regression model replaces convolution and pooling layers with fully connected layers in the dense block. Meanwhile, we maintain Batch Normalization layers from the original DenseNet to our novel networks. Batch Normalization is a typical regularization method with the advantage of accelerating the training process, reducing the impact of parameters scale, and allowing the utilization of higher learning rates \citep{j4ioffe2015batch}. 

The fundamental component of the DenseNet regression algorithm is the building block in \autoref{fig:label2}. There are three fully connected layers in this block, and each layer contains three operations, including batch normalization, dense, and ReLU activation function. Activation function is indispensable for the design of neural networks. Rectified linear unit (ReLU) activation function shown in \autoref{eq1} is widely used in neural networks, as it does not activate all the neurons at the same time, immensely reducing the computation \citep{20hanin2017approximating}. This is particularly beneficial to DenseNet regression where there is plenty of parameters to optimize. ReLU function is shown as follows
\begin{equation}\label{eq1}
  f(x)=\max(0,1)
\end{equation}
Another vital element in the building block is the concatenation shortcut, which is designed to append the input feature to the end of the output sequence in a building block. Furthermore, to satisfy the nonlinear regression tasks, a linear activation function is applied for the output (or top) layer. The number of layers in a building block is a hyperparameter and is determined by the accuracy and efficiency of neural networks. If there are few layers in the building block, the neural network would be too shallow and simple. Therefore, it is difficult to approximate nonlinear relationships. Nevertheless, if there are more layers in the building block, neural network parameters would be too large to optimize, which decreases the computation efficiency significantly. Meanwhile, under the same depth, neural networks with more layers in the building block have less concatenation
shortcuts, decreasing the efficiency of feature reuse. Hence the number of layers in the building block should be neither too small nor too large. Enlightened by the idea that there are three layers in each identity block and dense block in the optimal residual regression model \citep{dd7chen2020deep}, we make the input also go through three fully connected layers before concatenating in the building block. 

\autoref{fig:label3} shows an example of the architecture of a DenseNet regression model with four building blocks. The model begins with an input layer and is followed by four building blocks. Every building block is connected by concatenation shortcuts. The output layer is positioned in the end. Thus the total layers of this model is 13. The feature reuse is reflected in the concatenation shortcuts or curve arrows in \autoref{fig:label3}. The features of initial inputs and outputs of each building block would be transmitted to the following output layer through concatenation shortcuts. For example, the input of building block 2 not only contains the output of building block 1, but also includes the initial input features. By analogy, the input of top layer contains initial input features and the outputs of building block 1, 2, 3, and 4. Accordingly, the architecture of this model could keep feature reuse and enhance the performance of neural network on nonlinear regression. It should be noted that the specific number of building block needs to be optimized under given circumstances.

\section{Simulated data generation}\label{section3}
\begin{figure}[ht]
\centering
\includegraphics[width=8.8cm,height=8.13cm,scale=1]{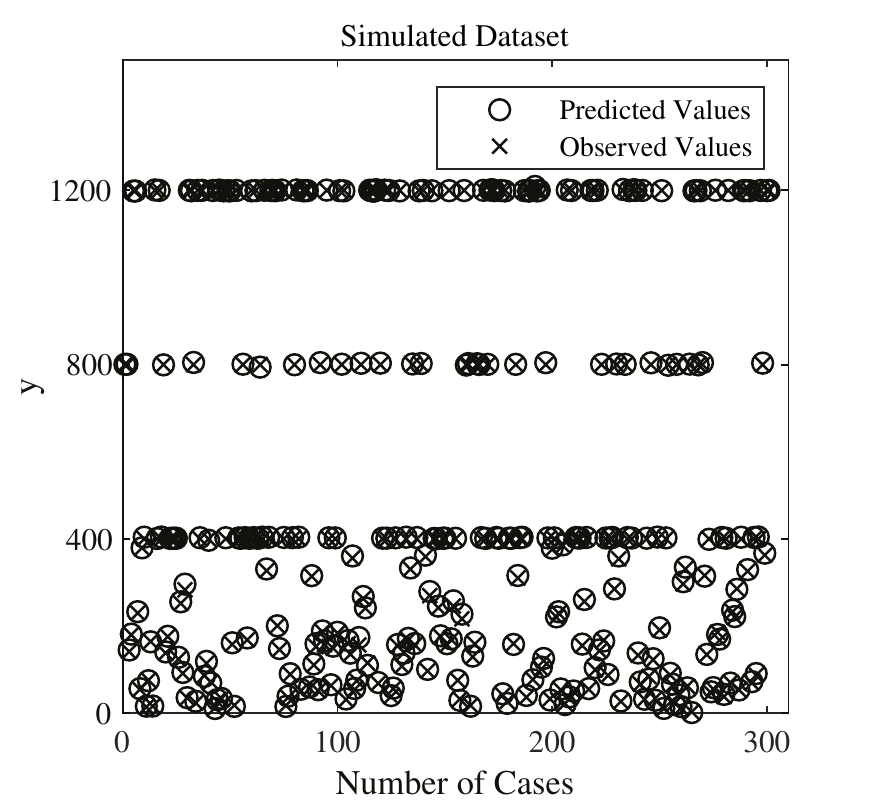}
\caption{Three hundred cases of simulated Dataset for DenseNet nonlinear regression model. $\bigcirc$ stands for the predicted values of optimal DenseNet regression model, and $\times$ stands for the true (observed) values of the simulated dataset.}
\label{fig:label4}
\end{figure}

To better understand the performance of this novel regression algorithm before it is applied to specific fields, a simulated nonlinear dataset is introduced to test the algorithm. The generation of this dataset is motivated and enlightened by the parameterization of fraction of cloud cover in numerical weather prediction, where people try to find a nonlinear relationship between meteorology variables and the fraction of cloud cover. The accurate prediction of fraction of cloud cover is critical to the success of solar energy power prediction. Fraction of cloud cover is defined as the proportion of a grid box covered by cloud, and thus the maximum fraction of cloud cover is 100$\%$. Following the similar data structure, we set a maximum value of 1200 in the simulated dataset. Moreover, in order to enhance the degree of nonlinearity, we add two smaller values (400 and 800) in the dataset, and finally, we have a nonlinear piecewise function as shown in \autoref{eq2}. In this scenario, 10,000,000 samples are generated from \autoref{eq2} where $x_i$ is uniformly distributed in the interval $[0, 4]$, i.e. $\left\{x_i\right\}_{i=0}^6\sim\ U[0,4]$.
\begin{equation}\label{eq2}
  y=\begin{cases}
   \sum\limits_{i=0}^6 (x_i)^i, & \text{if} \sum\limits_{i=0}^6 (x_i)^i<400;\\400, &\text{if}\ 400\le\sum\limits_{i=0}^6 (x_i)^i<800;\\800,&\text{if}\ 800\le\sum\limits_{i=0}^6 (x_i)^i<1200;\\1200,&\text{if}\sum\limits_{i=0}^6 (x_i)^i\ge1200.
    \end{cases}
\end{equation}
Three hundred cases from the generated dataset are shown in \autoref{fig:label4}. In this work, 6,750,000 samples are employed for training the DenseNet nonlinear regression model, while 750,000 samples served as validation data. The remaining 2,500,000 samples are utilized for testing.

\section{Results and discussion}\label{section4}
\subsection{DenseNet regression model specification}
Before training, the original data generated by \autoref{eq2} are standardized via the Min-Max scaler

\begin{equation}\label{eq3}
 \hat{u}_k=\dfrac{\max\limits_{i}u_i-u_k}{\max\limits_{i}u_i-\min\limits_{i}u_i}
\end{equation}
where $u_k$ is one of the sample data, $\hat{u}_k$stands for standardized data of $u_k$, and $i$ varies from one to the vector length.

Turning to the DenseNet regression model coding, Keras is employed as the application programming interface with TensorFlow as the backend. The computing environment is equipped with 10 CPUs and 10GB RAM. The type of CPUs is Intel(R) Xeon(R) CPU E5-2640 V4 with 2.40GH, and each core has two threads. Additionally, the Tesla P100-PCIE GPU is used for data training acceleration. It is produced by NVIDIA and has an 11.9GB memory.

For the model optimization, mean squared error (MSE) is served as the loss function as shown in \autoref{eq4}. To minimize the loss function, Adam method is applied in this work. Adam method computes individual adaptive learning rates for different parameters and integrates the merits of AdaGrad and RMSProp method, which work well in sparse gradients and online and non-stationary settings, respectively  \citep{kingma2014adam,duchi2011adaptive,reddi2018convergence}. The default learning rate of Adam method is 0.001 in Keras, but in practice, we notice that the validation loss oscillates in training. Therefore, the learning rate is set to 0.0001 in this section, so that the neural network has a better convergence performance. 
\begin{equation}\label{eq4}
 \text{Loss}=\dfrac{1}{N}\sum\limits_{m=1}^N (y_m-\hat{y}_m)^2
\end{equation}

Overfitting is a common issue when training machine learning models. It is probable that the loss gradually decreases during the training period while it rises in validation and testing. To prevent this and obtain a better model, early stopping strategy is used so that the neural network is very close to the epoch where the minimum validation loss arises. This strategy is effective and straightforward so that it is a popular regularization method in deep learning \citep{goodfellow2016deep}. After adding this strategy to the neural networks, the algorithm would stop when no progress has been made over the best-recorded validation loss for some pre-speciﬁed number (or patience) of epochs.

Primary parameter settings for the numerical validation scenario in this paper are as follows. The epoch number in training is 800, and the patience of early stopping is 100 epochs. The batch size for gradient descent is 5000 . And as mentioned above, the learning rate of Adam method is 0.0001. The magnitude of training loss, validation loss, and testing loss is $10^{-4}$. 

\begin{table*}[width=14.8cm]
\centering
\caption{Performance of DenseNet regression models with different depths.}\label{tab:table1}
\begin{threeparttable}
\begin{tabular}{llllllll}
\hline
\begin{tabular}[c]{@{}l@{}}\textbf{Depth of }\\\textbf{DenseNet}\end{tabular} & \begin{tabular}[c]{@{}l@{}}\textbf{Number of }\\\textbf{Parameters}\end{tabular} & \begin{tabular}[c]{@{}l@{}}\textbf{Expected}\\\textbf{Time}\end{tabular} & \begin{tabular}[c]{@{}l@{}}\textbf{Stopping }\\\textbf{Epoch}\end{tabular} & \begin{tabular}[c]{@{}l@{}}\textbf{Training}\\\textbf{Time}\end{tabular} & \begin{tabular}[c]{@{}l@{}}\textbf{Training }\\\textbf{Loss (10\textsuperscript{-4})}\end{tabular} & \begin{tabular}[c]{@{}l@{}}\textbf{Validation}\\\textbf{Loss (10\textsuperscript{-4})}\end{tabular} & \begin{tabular}[c]{@{}l@{}}\textbf{Testing}\\\textbf{Loss (10\textsuperscript{-4})}\end{tabular}  \\ 
\hline
4                                                                             & 407                                                                           & 00:50:00                                                                  & 800                                                                         & 00:50:37                                                                  & 9.5808                                                                                             & 6.5525                                                                                               & 6.6064                                                                                             \\
7                                                                             & 1,275                                                                         & 01:10:00                                                                  & 800                                                                         & 01:07:09                                                                  & 5.9951                                                                                             & 3.5137                                                                                               & 3.4935                                                                                             \\
10                                                                            & 4,187                                                                         & 01:30:00                                                                  & 800                                                                         & 01:26:16                                                                  & 3.5369                                                                                             & 2.4552                                                                                               & 2.5214                                                                                             \\
13                                                                            & 14,715                                                                        & 01:50:00                                                                  & 800                                                                         & 01:49:37                                                                  & 2.9436                                                                                             & 1.6492                                                                                               & 1.6349                                                                                             \\
16                                                                            & 54,587                                                                        & 02:30:00                                                                  & 607                                                                         & 01:52:02                                                                  & 2.3569                                                                                             & 1.8128                                                                                               & 1.7606                                                                                             \\
\textbf{19\tnote{*}}                                                                  & \textbf{209,595}                                                              & \textbf{03:40:00}                                                         & \textbf{800}                                                                & \textbf{03:36:56}                                                         & \textbf{1.9152}                                                                                    & \textbf{1.4785}                                                                                      & \textbf{1.5194}                                                                                    \\
22                                                                            & 820,667                                                                       & 06:10:00                                                                  & 329                                                                         & 02:31:16                                                                  & 2.6473                                                                                             & 2.0147                                                                                               & 2.0633                                                                                             \\
25                                                                            & 3,247,035                                                                     & 11:50:00                                                                  & 404                                                                         & 05:58:19                                                                  & 2.2012                                                                                             & 2.1022                                                                                               & 2.1704                                                                                             \\
28                                                                            & 12,916,667                                                                    & 28:20:00                                                                  & 333                                                                         & 11:48:34                                                                  & 2.3818                                                                                             & 3.4843                                                                                               & 3.4365                                                                                             \\
37                                                                            & 822,656,955                                                                   & OOM                                                                       & OOM                                                                         & OOM                                                                       & OOM                                                                                                & OOM                                                                                                  & OOM                                                                                                \\
\hline
\end{tabular}
\begin{tablenotes}
        \footnotesize
        \item[*] The optimal depth has a minimum testing loss. OOM stands for Out of Memory.
      \end{tablenotes}
\end{threeparttable}
\end{table*}

\subsection{DenseNet regression with the optimal depth}\label{section4.2}

To evaluate the effect of depth and find its optimal value, multiple DenseNet regression models with different depths ranging from 4 to 37 are trained in this part. The corresponding training parameters and performance of different regression models are shown in \autoref{tab:table1}. It is observed that with the rise of depth, the number of parameters and expected running time increase synchronously. When the depth is four, the DenseNet regression model has a high testing loss of $6.6064 \times 10^{-4}$, which shows that the neural network regression model with four layers is too simple to address complex and nonlinear regression tasks. As the depth of DenseNet regression model goes up, the testing loss goes down gradually. Remarkably, the testing loss reaches the lowest point, $1.5194 \times 10^{-4}$, when the depth is 19. Moreover, as the depth exceeds 19 and continues to increase, the testing loss goes up again. For the model with a depth of 37, the training result shows OOM (stands for Out of Memory) due to the tremendous number of parameters and computation. The outcome indicates that the depth of DenseNet regression model should be neither too small nor too large. In conclusion, the DenseNet regression model with depth 19 has the minimum testing loss and the best performance on simulated data. Therefore, the depth of optimal DenseNet regression model is 19, and we recommend employing this depth in the application part. It is also worth noticing that the testing loss has no significant difference as the depth ranges from 13 to 19. Hence if the computing resources are limited, we recommend setting the value of depth in the interval $[13,19]$ to get a compromise and balance between computing efficiency and accuracy.

\subsection{Comparisons with the baseline}
In this part, to evaluate the optimal DenseNet regression model, we also consider other regression techniques as the baseline on the dataset generated by \autoref{eq2}. The dataset is the same as \autoref{section3}. 10,000,000 samples are generated by \autoref{eq2} with 6,750,000 samples for training, 750,000 samples for validation and 2,500,000 samples for testing. Four linear models, including linear regression, ridge regression, lasso regression, and elastic regression, are applied to the simulated dataset. Nonlinear regression techniques incorporate conventional machine learning methods and artificial neural networks (ANNs), such as decision tree, support vector regression (SVR) machine, and deep residual regression. Neural networks contain the deep residual regression model and ANN Not Concatenated model. The residual regression model is a variant of ResNet. It replaces convolution and pooling layers with fully connected layers and has a good performance on nonlinear regression \citep{dd7chen2020deep}. ANN Not Concatenated model has the same structure as the optimal DenseNet regression model but has no concatenation shortcuts. This means that the depth of ANN not Concatenated model is 19. The epoch number and the patience of early stopping for neural networks are 800 and 100, respectively. The computing environment of residual regression model and ANN Not concatenated model is the same as DenseNet regression, modeling by Keras with TensorFlow as the backend. The four linear models, SVR model, and decision tree regression are built in Python with the scikit-learn package. The training results of different regression techniques are displayed in \autoref{tab:table2}, including the comparison items, training time, validation loss, and testing loss. \autoref{tab:table3} lists the optimal hyperparameters of all regression models mentioned above. Grid search method is employed to optimize the value of hyperparameters of each regression model, excluding linear regression and the last three neural networks. Usually, it takes a long time to train a support vector regression model on a large dataset. Therefore, the SVR model has been pre-trained before grid search. The pre-training gives smaller ranges of hyperparameters and thus improves the computing efficiency. Particularly, since the linear regression has no hyperparameters, the corresponding parameters are set by NA (NA stands for Not Applicable in \autoref{tab:table2} and \autoref{tab:table3}).

\begin{table*}[!htbp,pos=!ht,width=13.5cm]
\centering
\caption{\label{tab:test}Comparisons of DenseNet regression with the baseline models.}\label{tab:table2}
\arrayrulecolor{black}
\begin{threeparttable}
\begin{tabular}{llllll} 
\hline
\begin{tabular}[c]{@{}l@{}}\textbf{Regression}\\\textbf{Techniques Used }\end{tabular} & \begin{tabular}[c]{@{}l@{}}\textbf{Stopping }\\\textbf{Epoch }\end{tabular} & \begin{tabular}[c]{@{}l@{}}\textbf{Training}\\\textbf{Time }\end{tabular} & \begin{tabular}[c]{@{}l@{}}\textbf{Training}\\\textbf{Loss (10\textsuperscript{-4}) }\end{tabular} & \begin{tabular}[c]{@{}l@{}}\textbf{Validation}\\\textbf{Loss (10\textsuperscript{-4}) }\end{tabular} & \begin{tabular}[c]{@{}l@{}}\textbf{Testing}\\\textbf{Loss (10\textsuperscript{-4}) }\end{tabular}  \\ 
\hline
Linear regression                                                                       & NA                                                                           & 00:00:14                                                                   & 371.66                                                                                              & NA                                                                                                   & 371.52                                                                                              \\
Ridge regression                                                                        & NA                                                                           & 00:00:28                                                                   & 371.65                                                                                              & NA                                                                                                    & 371.55                                                                                              \\
Lasso regression                                                                        & NA                                                                           & 00:00:33                                                                   & 371.54                                                                                              & NA                                                                                                    & 371.87                                                                                              \\
Elastic regression                                                                      & NA                                                                           & 00:29:57                                                                   & 371.57                                                                                              & NA                                                                                                    & 371.80                                                                                              \\
Support Vector Regression                                                               & NA                                                                           & 03:09:09                                                                   & 142.85                                                                                              & NA                                                                                                    & 143.01                                                                                              \\
Decision tree regression                                                                & NA                                                                           & 00:16:05                                                                   & 2.8724                                                                                              & NA                                                                                                    & 4.7552                                                                                              \\
ANN Not Concatenated                                                                    & 800                                                                          & 02:11:14                                                                   & 6.5255                                                                                              & 4.0270                                                                                                & 3.9924                                                                                              \\
Residual regression                                                                     & 687                                                                          & 02:39:22                                                                   & 2.3889                                                                                              & 2.5674                                                                                                & 2.5931                                                                                              \\
\textbf{DenseNet regression\tnote{*}}                                                       & \textbf{800}                                                               & \textbf{03:36:56 }                                                        & \textbf{1.9152 }                                                                                   & \textbf{1.4785 }                                                                                     & \textbf{1.5194 }                                                                                   \\
\hline
\end{tabular}
\arrayrulecolor{black}
\begin{tablenotes}
        \footnotesize
        \item[*] DenseNet regression has a minimum testing loss. NA stands for Not Applicable.
      \end{tablenotes}
\end{threeparttable}
\end{table*}

The comparison results could be intuitively observed from the list of testing loss in \autoref{tab:table2}. Generally, the first four linear models have more significant testing loss than nonlinear models, reaching the magnitude of $10^{-1}$, which turns out that these four linear techniques are not applicable for nonlinear regression tasks. Among the remaining nonlinear regression techniques, the testing loss of support vector regression (SVR) machine is also with the magnitude of $10^{-1}$. Although support vector regression has been pre-trained before grid search, it still has a significant testing loss, which shows that support vector regression is not suitable for large datasets. The artificial neural network (ANN) without dense concatenation shortcuts in \autoref{tab:table2} has the same depth as the optimal DenseNet regression model but has a higher testing loss compared to the optimal DenseNet regression. It is also observed that DenseNet regression presents the best behavior among all the regression models in \autoref{tab:table2}, with the lowest testing loss ($1.5194 \times 10^{-4}$). This shows that concatenation shortcuts could enable feature reuse and keep critical information, and thus have a critical effect on the performance of DenseNet regression. Furthermore, one could note that the residual regression model has a second-smallest testing loss ($2.5931 \times 10^{-4}$) behind the DenseNet regression model. This is because the addition shortcuts in residual regression enable data flow and thus makes the model has a better performance. However, although residual regression has bypass addition shortcuts, the identity blocks and dense blocks are not densely connected. This illustrates that the outperformance of DenseNet regression to residual regression is due to the concatenation shortcuts and dense connection. Therefore, one could conclude that the topology of neural network and the way of connection (addition or concatenation) are essential to regression success. In conclusion, DenseNet regression is suitable for tackling nonlinear regression problems with high accuracy.

\begin{table*}[width=16.7cm]
\centering
\caption{Optimal hyperparameters of regression techniques.}\label{tab:table3}
\arrayrulecolor{black}
\begin{tabular}{llll} 
\arrayrulecolor{black}\hline
\begin{tabular}[c]{@{}l@{}}\textbf{Regression}\\\textbf{Techniques Used}\end{tabular} & \begin{tabular}[c]{@{}l@{}}\textbf{Name of }\\\textbf{Hyperparameters}\end{tabular}                                                                                  & \begin{tabular}[c]{@{}l@{}}\textbf{Range of}\\\textbf{Hyperparameters}\end{tabular}                                                                      & \begin{tabular}[c]{@{}l@{}}\textbf{Optimal }\\\textbf{Hyperparameters}\end{tabular}  \\ 
\arrayrulecolor{black}\hline
Linear regression                                                                     & NA                                                                                                                                                                   & NA                                                                                                                                                       & NA                                                                                   \\
Ridge regression                                                                      & Penalty parameter ~of L\textsuperscript{2}\textsuperscript{}                    ~norm $\alpha$;                                                                      & 10\textsuperscript{-10,}10\textsuperscript{-9}, 10\textsuperscript{-8} …, 10\textsuperscript{9}, 10\textsuperscript{10};                                 & 10\textsuperscript{2}                                                                \\
Lasso regression                                                                      & Penalty parameter ~of L\textsuperscript{1}  ~norm$\alpha$ ;                                                                                                          & 10\textsuperscript{-10},10\textsuperscript{-9}, 10\textsuperscript{-8} …, 10\textsuperscript{9}, 10\textsuperscript{10};                                 & 10\textsuperscript{-5}                                                               \\
Elastic regression                                                                    & \begin{tabular}[c]{@{}l@{}}Penalty parameter $\alpha$; \\Ratio of L\textsuperscript{1}~~~norm~$\rho$~;\end{tabular}                                                  & \begin{tabular}[c]{@{}l@{}}10\textsuperscript{-10},10\textsuperscript{-9},…, 10\textsuperscript{10};\\0.0, 0.1,…,0.9,1.0;\end{tabular}                   & \begin{tabular}[c]{@{}l@{}}10\textsuperscript{-5};\\1.0;\end{tabular}                \\
Support Vector Regression                                                             & \begin{tabular}[c]{@{}l@{}}Penalty parameter $\alpha$~;~ \\RBF kernel parameter~$\gamma$ ;\\Epsilon-tube parameter~$\epsilon$ ;\\Maximum iteration $N$;\end{tabular} & \begin{tabular}[c]{@{}l@{}}10, 10\textsuperscript{2}, 10\textsuperscript{3};\\10\textsuperscript{0},10, 10\textsuperscript{2};\\0.1;\\3000;\end{tabular} & \begin{tabular}[c]{@{}l@{}}10;\\10;\\0.1;\\3000;\end{tabular}                        \\
Decision tree regression                                                              & Maximum depth;                                                                                                                                                       & 1, 2, …,13,14;                                                                                                                                           & 14                                                                                   \\
ANN Not Concatenated                                                                  & Depth                                                                                                                                                                & NA                                                                                                                                                       & 19                                                                                   \\
Residual regression                                                                   & Width; Depth;                                                                                                                                                        & NA                                                                                                                                                       & 16; 28;                                                                              \\
DenseNet regression                                                                   & Depth                                                                                                                                                                & NA                                                                                                                                                       & 19                                                                                   \\
\hline
\end{tabular}
\arrayrulecolor{black}
\end{table*}

\subsection{The effect of input dimension}
It is known that the number of parameters in a deep neural network is related to input dimensions. The number of parameters becomes more remarkable as the neural network gets more input variables, and thus it takes more time for the algorithm to optimize and obtain a good validation. It is also noted that the computation progress even has OOM (Out of Memory) errors when the magnitude of parameters reaches $10^8$ under the given computation environment and simulated dataset, as shown in the depth optimization part of \autoref{section4.2}. Therefore, to enhance the computational efficiency, the input dimension should be limited. \autoref{tab:table4} lists the input dimensions of the optimal DenseNet regression model and the corresponding number of parameters. The optimal depth of used neural network here is 19. In \autoref{tab:table4}, the magnitude of parameter number reaches $10^5$ when the input dimension is 5. However, there is a sharp increase reaching 1,663,801 in the number of parameters as the input dimension extends to 20. Dramatically, when the input dimension is 50, the magnitude of parameter numbers becomes $10^7$. Nevertheless, when the input dimension varies from 50 to 100, the magnitude of parameter numbers does not change and is still kept at $10^7$. 

In conclusion, as the input dimension increases, the number of parameters goes up. Especially when the input dimension exceeds 80, the number of parameters increases sharply. If the input dimension reaches 200, the magnitude of parameter number would be $10^8$. In contrast, according to \autoref{tab:table1}, the outcome shows the out of memory (OOM) error if the magnitude of parameters reaches $10^8$. For the sake of computational efficiency, we conservatively suggest that if the computing capacity is as limited as our work, the input dimension for the optimal DenseNet regression model with depth 19 should be under 200 to avoid out of memory errors.

\begin{table}[pos=!ht,width=4.2cm]
\centering
\caption{The effect of input shape.}\label{tab:table4}
\arrayrulecolor{black}
\begin{tabular}{ll} 
\hline
\begin{tabular}[c]{@{}l@{}}\textbf{Input }\\\textbf{dimension  }\end{tabular} & \begin{tabular}[c]{@{}l@{}}\textbf{Number of }\\\textbf{parameters  }\end{tabular}  \\ 
\hline
5                                                                              & 108,751                                                                              \\
10                                                                             & 422,301                                                                              \\
15                                                                             & 940,651                                                                              \\
20                                                                             & 1,663,801                                                                            \\
30                                                                             & 3,724,501                                                                            \\
40                                                                             & 6,604,401                                                                            \\
45                                                                             & 8,351,551                                                                            \\
50                                                                             & 10,303,501                                                                           \\
60                                                                             & 14,821,801                                                                           \\
70                                                                             & 20,159,301                                                                           \\
80                                                                             & 26,316,001                                                                           \\
100                                                                            & 41,087,001                                                                           \\
150                                                                            & 92,350,501                                                                           \\
200                                                                            & 164,094,001                                                                          \\
\hline
\end{tabular}
\arrayrulecolor{black}
\end{table}

\section{Application of DenseNet regression on climate modeling }\label{section5}
Relative humidity is defined as the ratio of the water vapor pressure to the saturated water vapor pressure at a given temperature. It has a critical effect on cloud microphysics and dynamics, and hence plays a vital role in environment and climate \citep{fan2007effects}. However, the relative humidity is not accurate under vapor supersaturation circumstances since the formation of cloud condensation nuclei needs water vapor to be supersaturated in the air, and there is no widely accepted and reliable method to measure the supersaturated vapor pressure accurately at present \citep{shen2018method}. To address this issue, this section uses the DenseNet regression model with optimal depth to quantify the nonlinear relationship between relative humidity and other environmental factors. 	

\begin{figure}[ht]
\centering
\includegraphics[width=8.8cm,height=8.13cm,scale=1]{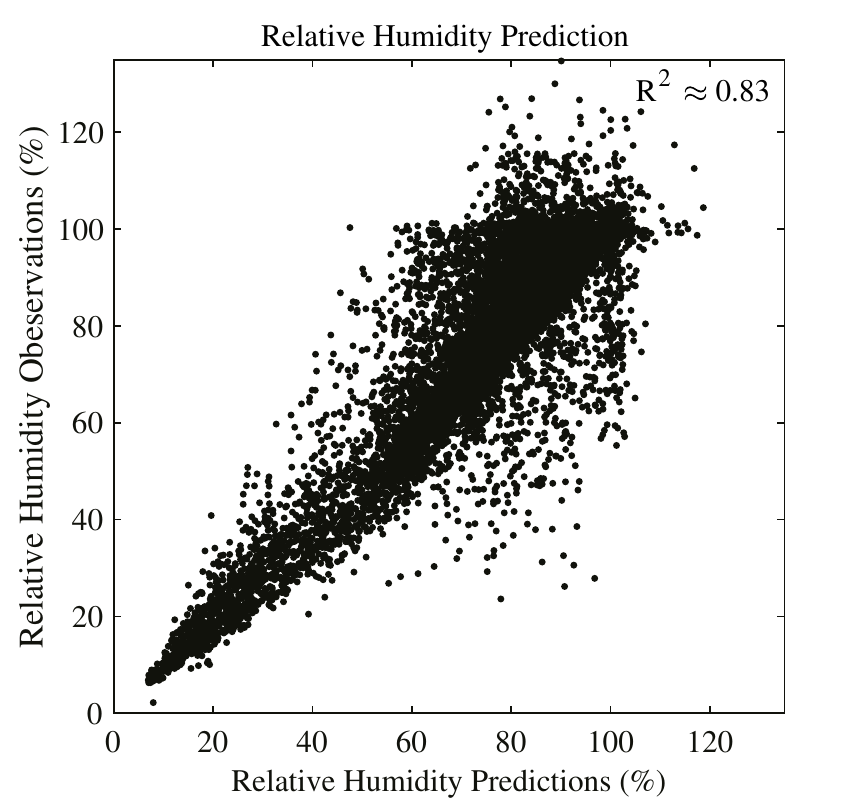}
\caption{Relative Humidity prediction using the optimal DenseNet regression model. The figure shows the first 20000 cases of relative humidity prediction. The correlation coefficient $\rho$ is approximately 0.91 and the corresponding determination coefficient $R^2$ is near 0.83. This shows that DenseNet regression has high performance under real-world scenarios.}
\label{fig:label5}
\end{figure}

Our data is from EAR5 hourly reanalysis datasets on the 1000 hPa pressure level of ECMEF (European Centre for Medium-Range Weather Forecasts) \citep{hersbach2018era5}. The input features of DenseNet regression are temperature and specific humidity. The response variable is relative humidity. The dataset is selected from 00:00:00 a.m. to 23:00:00 p.m. on Sep. 1st, 2007 with a spatial resolution of $0.25^{\circ} \times 0.25^{\circ}$. There are 24,917,709 samples in total. The optimal DenseNet regression model is trained on 16,819,454 samples and validated on 1,868,828 samples. The remaining 6,229,427 samples are testing data. The batch size is 20,000, and the setting of other parameters are the same as the optimal DenseNet regression setting in \autoref{section4}. Furthermore, the consequence shows that the correlation coefficient ($\rho$) for the fitted values and observed values of testing data is $0.9095$, which is shown in \autoref{fig:label5}. The corresponding determination coefficient ($R^2$) is approximately 0.83. This indicates that the fitted values and observed ones are highly correlated. The average relative error in testing for relative humidity is $6.23\%$, verifying that the DenseNet regression model behaves excellently in practice. 

\section{Conclusion and future work}\label{section6}
In this paper, we develop a novel densely connected neural networks for nonlinear regression. The convolutional DenseNet behaves well in image processing. However, when it is applied to regression tasks, the independence of input features makes the convolutional neural network lose critical information from input variables. Therefore, it is not suitable for nonlinear regression. To address this issue, we replace convolutional layers and pooling layers with fully connected layers, and reserve the DenseNet dense concatenation connections to enhance feature reuse in the regression model. The new regression model is numerically evaluated on simulated data, and the results recommend an optimal depth (19) and input dimensions (under 200) for the regression model. In addition, we compare  DenseNet regression with other baseline techniques, such as support vector regression, decision tree regression, and deep residual regression. It turns out that the DenseNet regression model with optimal depth has the lowest testing loss. Finally, the optimal DenseNet regression is applied to predict relative humidity, and we obtain a low average relative error, which indicates that the DenseNet regression model is applicable in practice and could advance environmental science. 

In the future, we intend to apply the DenseNet regression model to the parameterization of subgrid-scale process of large eddy simulation of turbulence at the atmospheric boundary layer. In addition, we will also employ the DenseNet regression to estimate global terrestrial carbon fluxes using net ecosystem exchange (NEE), gross primary production (GPP), and ecosystem respiration (RECO) from FLUXNET2015 dataset.

% Use the style of numbering in square brackets.
% If nothing is used, default style will be taken.
%\begin{enumerate}[a)]
%\item 
%\item 
%\item 
%\end{enumerate}  

%\begin{itemize}
%\item 
%\item 
%\item 
%\end{itemize}  

%\begin{description}
%\item[]
%\item[] 
%\item[] 
%\end{description}  

% Figure
% \begin{figure}[<options>]
% 	\centering
% 		\includegraphics[<options>]{}
% 	  \caption{}\label{fig1}
% \end{figure}

% Uncomment and use as the case may be
%\begin{theorem} 
%\end{theorem}

% Uncomment and use as the case may be
%\begin{lemma} 
%\end{lemma}


\printcredits
\printFunding{This research receives no grant from any funding agency in the public, commercial, or not-for-profit sectors.}

\declareInterests{The authors declare that they have no known competing financial interests or personal relationships that could have appeared to influence the work reported in this paper.}

\printAcknowledge{Clemson University is acknowledged for generous allotment of computing time on Palmetto cluster. The authors thank all the anonymous reviewers for their constructive comments. The authors also thank all the editors for their careful proofreading.The code of optimal DenseNet regression model could be found at \url{https://github.com/DowellChan/DenseNetRegression}}


\newcommand{\DOI}[1]{doi: \href{https://doi.org/#1}{#1}}
\bibliographystyle{cas-model2-names}

\bibliography{cas-refs}

\begin{thebibliography}{43}
\expandafter\ifx\csname natexlab\endcsname\relax\def\natexlab#1{#1}\fi
\providecommand{\url}[1]{\texttt{#1}}
\providecommand{\href}[2]{#2}
\providecommand{\path}[1]{#1}
\providecommand{\DOIprefix}{doi:}
\providecommand{\ArXivprefix}{arXiv:}
\providecommand{\URLprefix}{URL: }
\providecommand{\Pubmedprefix}{pmid:}
\providecommand{\doi}[1]{\href{http://dx.doi.org/#1}{\path{#1}}}
\providecommand{\Pubmed}[1]{\href{pmid:#1}{\path{#1}}}
\providecommand{\bibinfo}[2]{#2}
\ifx\xfnm\relax \def\xfnm[#1]{\unskip,\space#1}\fi
%Type = Article
\bibitem[{Ahn et~al.(2012)Ahn, Byun, Oh and Kim}]{3ahn2012using}
\bibinfo{author}{Ahn, J.J.}, \bibinfo{author}{Byun, H.W.}, \bibinfo{author}{Oh,
  K.J.}, \bibinfo{author}{Kim, T.Y.}, \bibinfo{year}{2012}.
\newblock \bibinfo{title}{Using ridge regression with genetic algorithm to
  enhance real estate appraisal forecasting}.
\newblock \bibinfo{journal}{Expert Systems with Applications}
  \bibinfo{volume}{39}, \bibinfo{pages}{8369--8379}.
\newblock \bibinfo{note}{Doi:{\color{blue}
  \href{https://doi.org/10.1016/j.eswa.2012.01.183}{10.1016/j.eswa.2012.01.183}}}.
%Type = Article
\bibitem[{Chen et~al.(2020)Chen, Hu, Nian and Yang}]{dd7chen2020deep}
\bibinfo{author}{Chen, D.}, \bibinfo{author}{Hu, F.}, \bibinfo{author}{Nian,
  G.}, \bibinfo{author}{Yang, T.}, \bibinfo{year}{2020}.
\newblock \bibinfo{title}{Deep residual learning for nonlinear regression}.
\newblock \bibinfo{journal}{Entropy} \bibinfo{volume}{22},
  \bibinfo{pages}{193}.
\newblock \bibinfo{note}{Doi:{\color{blue}
  \href{https://doi.org/10.3390/e22020193}{10.3390/e22020193}}}.
%Type = Article
\bibitem[{Chui et~al.(1996)Chui, Li and Mhaskar}]{j5chui1996limitations}
\bibinfo{author}{Chui, C.K.}, \bibinfo{author}{Li, X.},
  \bibinfo{author}{Mhaskar, H.N.}, \bibinfo{year}{1996}.
\newblock \bibinfo{title}{Limitations of the approximation capabilities of
  neural networks with one hidden layer}.
\newblock \bibinfo{journal}{Advances in Computational Mathematics}
  \bibinfo{volume}{5}, \bibinfo{pages}{233--243}.
\newblock \bibinfo{note}{Doi:{\color{blue}
  \href{https://doi.org/10.1007/BF02124745}{10.1007/BF02124745}}}.
%Type = Article
\bibitem[{Danasingh et~al.(2020)Danasingh, Epiphany
  et~al.}]{danasingh2020identifying}
\bibinfo{author}{Danasingh, A.A.G.S.}, \bibinfo{author}{Epiphany, J.L.},
  et~al., \bibinfo{year}{2020}.
\newblock \bibinfo{title}{Identifying redundant features using unsupervised
  learning for high-dimensional data}.
\newblock \bibinfo{journal}{SN Applied Sciences} \bibinfo{volume}{2},
  \bibinfo{pages}{1--10}.
\newblock \bibinfo{note}{Doi:{\color{blue}
  \href{https://doi.org/10.1007/s42452-020-3157-6}{10.1007/s42452-020-3157-6}}}.
%Type = Article
\bibitem[{Duchi et~al.(2011)Duchi, Hazan and Singer}]{duchi2011adaptive}
\bibinfo{author}{Duchi, J.}, \bibinfo{author}{Hazan, E.},
  \bibinfo{author}{Singer, Y.}, \bibinfo{year}{2011}.
\newblock \bibinfo{title}{Adaptive subgradient methods for online learning and
  stochastic optimization.}
\newblock \bibinfo{journal}{Journal of machine learning research}
  \bibinfo{volume}{12}.
%Type = Article
\bibitem[{Fan et~al.(2007)Fan, Zhang, Li and Tao}]{fan2007effects}
\bibinfo{author}{Fan, J.}, \bibinfo{author}{Zhang, R.}, \bibinfo{author}{Li,
  G.}, \bibinfo{author}{Tao, W.K.}, \bibinfo{year}{2007}.
\newblock \bibinfo{title}{Effects of aerosols and relative humidity on cumulus
  clouds}.
\newblock \bibinfo{journal}{Journal of Geophysical Research: Atmospheres}
  \bibinfo{volume}{112}.
\newblock \bibinfo{note}{Doi:{\color{blue}
  \href{https://doi.org/10.1029/2006JD008136}{10.1029/2006JD008136}}}.
%Type = Article
\bibitem[{Funahashi(1989)}]{14funahashi1989approximate}
\bibinfo{author}{Funahashi, K.I.}, \bibinfo{year}{1989}.
\newblock \bibinfo{title}{On the approximate realization of continuous mappings
  by neural networks}.
\newblock \bibinfo{journal}{Neural networks} \bibinfo{volume}{2},
  \bibinfo{pages}{183--192}.
\newblock \bibinfo{note}{Doi:{\color{blue}
  \href{https://doi.org/10.1016/0893-6080(89)90003-8}{10.1016/0893-6080(89)90003-8}}}.
%Type = Article
\bibitem[{Gibert et~al.(2018)Gibert, Horsburgh, Athanasiadis and
  Holmes}]{j9gibert2018environmental}
\bibinfo{author}{Gibert, K.}, \bibinfo{author}{Horsburgh, J.S.},
  \bibinfo{author}{Athanasiadis, I.N.}, \bibinfo{author}{Holmes, G.},
  \bibinfo{year}{2018}.
\newblock \bibinfo{title}{Environmental data science}.
\newblock \bibinfo{journal}{Environmental Modelling \& Software}
  \bibinfo{volume}{106}, \bibinfo{pages}{4--12}.
\newblock \bibinfo{note}{Doi:{\color{blue}
  \href{https://doi.org/10.1016/j.envsoft.2018.04.005}{10.1016/j.envsoft.2018.04.005}}}.
%Type = Book
\bibitem[{Goodfellow et~al.(2016)Goodfellow, Bengio and
  Courville}]{goodfellow2016deep}
\bibinfo{author}{Goodfellow, I.}, \bibinfo{author}{Bengio, Y.},
  \bibinfo{author}{Courville, A.}, \bibinfo{year}{2016}.
\newblock \bibinfo{title}{Deep learning}.
\newblock \bibinfo{publisher}{MIT press}.
%Type = Article
\bibitem[{Hanin and Sellke(2017)}]{20hanin2017approximating}
\bibinfo{author}{Hanin, B.}, \bibinfo{author}{Sellke, M.},
  \bibinfo{year}{2017}.
\newblock \bibinfo{title}{Approximating continuous functions by relu nets of
  minimal width}.
\newblock \bibinfo{journal}{arXiv preprint arXiv:1710.11278} .
%Type = Misc
\bibitem[{Hersbach et~al.(2018)Hersbach, Bell, Berrisford, Biavati,
  Hor{\'a}nyi, Mu{\~n}oz~Sabater, Nicolas, Peubey, Radu, Rozum
  et~al.}]{hersbach2018era5}
\bibinfo{author}{Hersbach, H.}, \bibinfo{author}{Bell, B.},
  \bibinfo{author}{Berrisford, P.}, \bibinfo{author}{Biavati, G.},
  \bibinfo{author}{Hor{\'a}nyi, A.}, \bibinfo{author}{Mu{\~n}oz~Sabater, J.},
  \bibinfo{author}{Nicolas, J.}, \bibinfo{author}{Peubey, C.},
  \bibinfo{author}{Radu, R.}, \bibinfo{author}{Rozum, I.}, et~al.,
  \bibinfo{year}{2018}.
\newblock \bibinfo{title}{Era5 hourly data on pressure levels from 1979 to
  present, copernicus climate change service (c3s) climate data store (cds)}.
\newblock \bibinfo{note}{Doi:{\color{blue} \href{
  https://doi.org/10.24381/cds.bd0915c6}{10.24381/cds.bd0915c6}}}.
%Type = Article
\bibitem[{Hornik et~al.(1989)Hornik, Stinchcombe and
  White}]{13hornik1989multilayer}
\bibinfo{author}{Hornik, K.}, \bibinfo{author}{Stinchcombe, M.},
  \bibinfo{author}{White, H.}, \bibinfo{year}{1989}.
\newblock \bibinfo{title}{Multilayer feedforward networks are universal
  approximators}.
\newblock \bibinfo{journal}{Neural networks} \bibinfo{volume}{2},
  \bibinfo{pages}{359--366}.
\newblock \bibinfo{note}{Doi:{\color{blue}
  \href{https://doi.org/10.1016/0893-6080(89)90020-8}{10.1016/0893-6080(89)90020-8}}}.
%Type = Inproceedings
\bibitem[{Huang et~al.(2017)Huang, Liu, Van Der~Maaten and
  Weinberger}]{18huang2017densely}
\bibinfo{author}{Huang, G.}, \bibinfo{author}{Liu, Z.}, \bibinfo{author}{Van
  Der~Maaten, L.}, \bibinfo{author}{Weinberger, K.Q.}, \bibinfo{year}{2017}.
\newblock \bibinfo{title}{Densely connected convolutional networks}, in:
  \bibinfo{booktitle}{Proceedings of the IEEE conference on computer vision and
  pattern recognition}, pp. \bibinfo{pages}{4700--4708}.
\newblock \bibinfo{note}{Doi:{\color{blue}
  \href{https://doi.org/10.1109/CVPR.2017.243}{10.1109/CVPR.2017.243}}}.
%Type = Inproceedings
\bibitem[{Ioffe and Szegedy(2015)}]{j4ioffe2015batch}
\bibinfo{author}{Ioffe, S.}, \bibinfo{author}{Szegedy, C.},
  \bibinfo{year}{2015}.
\newblock \bibinfo{title}{Batch normalization: Accelerating deep network
  training by reducing internal covariate shift}, in:
  \bibinfo{booktitle}{International conference on machine learning},
  \bibinfo{organization}{PMLR}. pp. \bibinfo{pages}{448--456}.
\newblock \bibinfo{note}{Doi:{\color{blue}
  \href{https://doi.org/10.5555/3045118.3045167}{10.5555/3045118.3045167}}}.
%Type = Article
\bibitem[{Khaki and Wang(2019)}]{j7khaki2019crop}
\bibinfo{author}{Khaki, S.}, \bibinfo{author}{Wang, L.}, \bibinfo{year}{2019}.
\newblock \bibinfo{title}{Crop yield prediction using deep neural networks}.
\newblock \bibinfo{journal}{Frontiers in plant science} \bibinfo{volume}{10},
  \bibinfo{pages}{621}.
\newblock \bibinfo{note}{Doi:{\color{blue}
  \href{https://https://doi.org/10.3389/fpls.2019.00621}{10.3389/fpls.2019.00621}}}.
%Type = Article
\bibitem[{Kingma and Ba(2014)}]{kingma2014adam}
\bibinfo{author}{Kingma, D.P.}, \bibinfo{author}{Ba, J.}, \bibinfo{year}{2014}.
\newblock \bibinfo{title}{Adam: A method for stochastic optimization}.
\newblock \bibinfo{journal}{arXiv preprint arXiv:1412.6980} .
%Type = Article
\bibitem[{Krishna et~al.(2020)Krishna, Shanmugam and
  Nagamani}]{j12krishna2020multiparametric}
\bibinfo{author}{Krishna, K.V.}, \bibinfo{author}{Shanmugam, P.},
  \bibinfo{author}{Nagamani, P.V.}, \bibinfo{year}{2020}.
\newblock \bibinfo{title}{A multiparametric nonlinear regression approach for
  the estimation of global surface ocean pco 2 using satellite oceanographic
  data}.
\newblock \bibinfo{journal}{IEEE Journal of Selected Topics in Applied Earth
  Observations and Remote Sensing} \bibinfo{volume}{13},
  \bibinfo{pages}{6220--6235}.
\newblock \bibinfo{note}{Doi:{\color{blue}
  \href{https://doi.org/10.1109/JSTARS.2020.3026363}{10.1109/JSTARS.2020.3026363}}}.
%Type = Article
\bibitem[{K{\r u}rkov{\'a}(1992)}]{15kuurkova1992kolmogorov}
\bibinfo{author}{K{\r u}rkov{\'a}, V.}, \bibinfo{year}{1992}.
\newblock \bibinfo{title}{Kolmogorov's theorem and multilayer neural networks}.
\newblock \bibinfo{journal}{Neural networks} \bibinfo{volume}{5},
  \bibinfo{pages}{501--506}.
\newblock \bibinfo{note}{Doi:{\color{blue}
  \href{https://doi.org/10.1016/0893-6080(92)90012-8}{10.1016/0893-6080(92)90012-8}}}.
%Type = Article
\bibitem[{Lathuili{\`e}re et~al.(2019)Lathuili{\`e}re, Mesejo, Alameda-Pineda
  and Horaud}]{dd6lathuiliere2019comprehensive}
\bibinfo{author}{Lathuili{\`e}re, S.}, \bibinfo{author}{Mesejo, P.},
  \bibinfo{author}{Alameda-Pineda, X.}, \bibinfo{author}{Horaud, R.},
  \bibinfo{year}{2019}.
\newblock \bibinfo{title}{A comprehensive analysis of deep regression}.
\newblock \bibinfo{journal}{IEEE transactions on pattern analysis and machine
  intelligence} \bibinfo{volume}{42}, \bibinfo{pages}{2065--2081}.
\newblock \bibinfo{note}{Doi:{\color{blue}
  \href{https://doi.org/10.1109/TPAMI.2019.2910523}{10.1109/TPAMI.2019.2910523}}}.
%Type = Article
\bibitem[{Loh(2011)}]{11loh2011classification}
\bibinfo{author}{Loh, W.Y.}, \bibinfo{year}{2011}.
\newblock \bibinfo{title}{Classification and regression trees}.
\newblock \bibinfo{journal}{Wiley interdisciplinary reviews: data mining and
  knowledge discovery} \bibinfo{volume}{1}, \bibinfo{pages}{14--23}.
\newblock \bibinfo{note}{Doi:{\color{blue}
  \href{https://doi.org/10.1002/widm.8}{10.1002/widm.8}}}.
%Type = Article
\bibitem[{Majda and Harlim(2012)}]{7majda2012physics}
\bibinfo{author}{Majda, A.J.}, \bibinfo{author}{Harlim, J.},
  \bibinfo{year}{2012}.
\newblock \bibinfo{title}{Physics constrained nonlinear regression models for
  time series}.
\newblock \bibinfo{journal}{Nonlinearity} \bibinfo{volume}{26},
  \bibinfo{pages}{201}.
\newblock \bibinfo{note}{Doi:{\color{blue}
  \href{https://doi.org/10.1088/0951-7715/26/1/201}{10.1088/0951-7715/26/1/201}}}.
%Type = Article
\bibitem[{Ostertagov{\'a}(2012)}]{9ostertagova2012modelling}
\bibinfo{author}{Ostertagov{\'a}, E.}, \bibinfo{year}{2012}.
\newblock \bibinfo{title}{Modelling using polynomial regression}.
\newblock \bibinfo{journal}{Procedia Engineering} \bibinfo{volume}{48},
  \bibinfo{pages}{500--506}.
\newblock \bibinfo{note}{Doi:{\color{blue}
  \href{https://doi.org/10.1016/j.proeng.2012.09.545}{10.1016/j.proeng.2012.09.545}}}.
%Type = Article
\bibitem[{Pandey et~al.(2020)Pandey, Zakwan, Sharma and
  Ahmad}]{1pandey2020multiple}
\bibinfo{author}{Pandey, M.}, \bibinfo{author}{Zakwan, M.},
  \bibinfo{author}{Sharma, P.}, \bibinfo{author}{Ahmad, Z.},
  \bibinfo{year}{2020}.
\newblock \bibinfo{title}{Multiple linear regression and genetic algorithm
  approaches to predict temporal scour depth near circular pier in non-cohesive
  sediment}.
\newblock \bibinfo{journal}{ISH Journal of Hydraulic Engineering}
  \bibinfo{volume}{26}, \bibinfo{pages}{96--103}.
\newblock \bibinfo{note}{Doi:{\color{blue}
  \href{http://doi.org/10.1080/09715010.2018.1457455}{10.1080/09715010.2018.1457455}}}.
%Type = Inproceedings
\bibitem[{Reddi et~al.(2018)Reddi, Kale and Kumar}]{reddi2018convergence}
\bibinfo{author}{Reddi, S.J.}, \bibinfo{author}{Kale, S.},
  \bibinfo{author}{Kumar, S.}, \bibinfo{year}{2018}.
\newblock \bibinfo{title}{On the convergence of adam and beyond}, in:
  \bibinfo{booktitle}{International Conference on Learning Representations}.
%Type = Incollection
\bibitem[{Rehana(2019)}]{j11rehana2019river}
\bibinfo{author}{Rehana, S.}, \bibinfo{year}{2019}.
\newblock \bibinfo{title}{River water temperature modelling under climate
  change using support vector regression}, in: \bibinfo{booktitle}{Hydrology in
  a Changing World}. \bibinfo{publisher}{Springer}, pp.
  \bibinfo{pages}{171--183}.
\newblock \bibinfo{note}{Doi:{\color{blue}
  \href{https://doi.org/10.1007/978-3-030-02197-9_8}{10.1007/978-3-030-02197-9_8}}}.
%Type = Book
\bibitem[{Rhinehart(2016)}]{6rhinehart2016nonlinear}
\bibinfo{author}{Rhinehart, R.R.}, \bibinfo{year}{2016}.
\newblock \bibinfo{title}{Nonlinear regression modeling for engineering
  applications: modeling, model validation, and enabling design of
  experiments}.
\newblock \bibinfo{publisher}{John Wiley \& Sons}.
\newblock \bibinfo{note}{Doi:{\color{blue}
  \href{https://doi.org/10.1115/1.861NONQ}{10.1115/1.861NONQ}}}.
%Type = Inproceedings
\bibitem[{Rojek et~al.(2021)Rojek, Wyrzykowski and Gepner}]{dd1rojek2021ai}
\bibinfo{author}{Rojek, K.}, \bibinfo{author}{Wyrzykowski, R.},
  \bibinfo{author}{Gepner, P.}, \bibinfo{year}{2021}.
\newblock \bibinfo{title}{Ai-accelerated cfd simulation based on openfoam and
  cpu/gpu computing}, in: \bibinfo{booktitle}{International Conference on
  Computational Science}, \bibinfo{organization}{Springer}. pp.
  \bibinfo{pages}{373--385}.
\newblock \bibinfo{note}{Doi:{\color{blue}
  \href{https://doi.org/10.1007/978-3-030-77964-1_29}{10.1007/978-3-030-77964-1_29}}}.
%Type = Article
\bibitem[{Rosenlund et~al.(2008)Rosenlund, Forastiere, Stafoggia, Porta,
  Perucci, Ranzi, Nussio and Perucci}]{j10rosenlund2008comparison}
\bibinfo{author}{Rosenlund, M.}, \bibinfo{author}{Forastiere, F.},
  \bibinfo{author}{Stafoggia, M.}, \bibinfo{author}{Porta, D.},
  \bibinfo{author}{Perucci, M.}, \bibinfo{author}{Ranzi, A.},
  \bibinfo{author}{Nussio, F.}, \bibinfo{author}{Perucci, C.A.},
  \bibinfo{year}{2008}.
\newblock \bibinfo{title}{Comparison of regression models with land-use and
  emissions data to predict the spatial distribution of traffic-related air
  pollution in rome}.
\newblock \bibinfo{journal}{Journal of exposure science \& environmental
  epidemiology} \bibinfo{volume}{18}, \bibinfo{pages}{192--199}.
\newblock \bibinfo{note}{Doi:{\color{blue}
  \href{https://doi.org/10.1038/sj.jes.7500571}{10.1038/sj.jes.7500571}}}.
%Type = Inproceedings
\bibitem[{Saleh et~al.(2019)Saleh, Hossny and Nahavandi}]{jj3saleh2019real}
\bibinfo{author}{Saleh, K.}, \bibinfo{author}{Hossny, M.},
  \bibinfo{author}{Nahavandi, S.}, \bibinfo{year}{2019}.
\newblock \bibinfo{title}{Real-time intent prediction of pedestrians for
  autonomous ground vehicles via spatio-temporal densenet}, in:
  \bibinfo{booktitle}{2019 International Conference on Robotics and Automation
  (ICRA)}, \bibinfo{organization}{IEEE}. pp. \bibinfo{pages}{9704--9710}.
\newblock \bibinfo{note}{Doi:{\color{blue} \href{
  https://doi.org/10.1109/ICRA.2019.8793991}{10.1109/ICRA.2019.8793991}}}.
%Type = Article
\bibitem[{Scholz and J{\"u}ttler(2021)}]{dd3scholz2021parameterization}
\bibinfo{author}{Scholz, F.}, \bibinfo{author}{J{\"u}ttler, B.},
  \bibinfo{year}{2021}.
\newblock \bibinfo{title}{Parameterization for polynomial curve approximation
  via residual deep neural networks}.
\newblock \bibinfo{journal}{Computer Aided Geometric Design}
  \bibinfo{volume}{85}, \bibinfo{pages}{101977}.
\newblock \bibinfo{note}{Doi:{\color{blue}
  \href{https://doi.org/10.1016/j.cagd.2021.101977}{10.1016/j.cagd.2021.101977}}}.
%Type = Article
\bibitem[{Shen et~al.(2018)Shen, Zhao, Ma, Tao, Zhao, Yu and
  Kuang}]{shen2018method}
\bibinfo{author}{Shen, C.}, \bibinfo{author}{Zhao, C.}, \bibinfo{author}{Ma,
  N.}, \bibinfo{author}{Tao, J.}, \bibinfo{author}{Zhao, G.},
  \bibinfo{author}{Yu, Y.}, \bibinfo{author}{Kuang, Y.}, \bibinfo{year}{2018}.
\newblock \bibinfo{title}{Method to estimate water vapor supersaturation in the
  ambient activation process using aerosol and droplet measurement data}.
\newblock \bibinfo{journal}{Journal of Geophysical Research: Atmospheres}
  \bibinfo{volume}{123}, \bibinfo{pages}{10--606}.
\newblock \bibinfo{note}{Doi:{\color{blue} \href{
  https://doi.org/10.1029/2018JD028315}{10.1029/2018JD028315}}}.
%Type = Inproceedings
\bibitem[{Shi et~al.(2020)Shi, Copot and Vanlanduit}]{dd4shi2020deep}
\bibinfo{author}{Shi, L.}, \bibinfo{author}{Copot, C.},
  \bibinfo{author}{Vanlanduit, S.}, \bibinfo{year}{2020}.
\newblock \bibinfo{title}{A deep regression model for safety control in visual
  servoing applications}, in: \bibinfo{booktitle}{2020 Fourth IEEE
  International Conference on Robotic Computing (IRC)},
  \bibinfo{organization}{IEEE}. pp. \bibinfo{pages}{360--366}.
\newblock \bibinfo{note}{Doi:{\color{blue}
  \href{https://doi.org/10.1109/IRC.2020.00063}{10.1109/IRC.2020.00063}}}.
%Type = Article
\bibitem[{Shi et~al.(2021)Shi, Copot and Vanlanduit}]{dd5shi2021bayesian}
\bibinfo{author}{Shi, L.}, \bibinfo{author}{Copot, C.},
  \bibinfo{author}{Vanlanduit, S.}, \bibinfo{year}{2021}.
\newblock \bibinfo{title}{A bayesian deep neural network for safe visual
  servoing in human--robot interaction}.
\newblock \bibinfo{journal}{Frontiers in Robotics and AI} \bibinfo{volume}{8}.
\newblock \bibinfo{note}{Doi:{\color{blue}
  \href{https://doi.org/10.3389/frobt.2021.687031}{10.3389/frobt.2021.687031}}}.
%Type = Article
\bibitem[{Shin et~al.(2021)Shin, Ge, Lampmann and Pfitzner}]{dd2shin2021data}
\bibinfo{author}{Shin, J.}, \bibinfo{author}{Ge, Y.},
  \bibinfo{author}{Lampmann, A.}, \bibinfo{author}{Pfitzner, M.},
  \bibinfo{year}{2021}.
\newblock \bibinfo{title}{A data-driven subgrid scale model in large eddy
  simulation of turbulent premixed combustion}.
\newblock \bibinfo{journal}{Combustion and Flame} \bibinfo{volume}{231},
  \bibinfo{pages}{111486}.
\newblock \bibinfo{note}{Doi:{\color{blue}
  \href{https://doi.org/10.1016/j.combustflame.2021.111486}{10.1016/j.combustflame.2021.111486}}}.
%Type = Article
\bibitem[{Smola and Sch{\"o}lkopf(2004)}]{10smola2004tutorial}
\bibinfo{author}{Smola, A.J.}, \bibinfo{author}{Sch{\"o}lkopf, B.},
  \bibinfo{year}{2004}.
\newblock \bibinfo{title}{A tutorial on support vector regression}.
\newblock \bibinfo{journal}{Statistics and computing} \bibinfo{volume}{14},
  \bibinfo{pages}{199--222}.
\newblock \bibinfo{note}{Doi:{\color{blue}
  \href{https://doi.org/10.1023/B:STCO.0000035301.49549.88}{10.1023/B:STCO.0000035301.49549.88}}}.
%Type = Article
\bibitem[{Sun et~al.(2018)Sun, Gui, Li, Liu and An}]{jj1sun2018resinnet}
\bibinfo{author}{Sun, X.}, \bibinfo{author}{Gui, G.}, \bibinfo{author}{Li, Y.},
  \bibinfo{author}{Liu, R.P.}, \bibinfo{author}{An, Y.}, \bibinfo{year}{2018}.
\newblock \bibinfo{title}{Resinnet: A novel deep neural network with feature
  reuse for internet of things}.
\newblock \bibinfo{journal}{IEEE Internet of Things Journal}
  \bibinfo{volume}{6}, \bibinfo{pages}{679--691}.
\newblock \bibinfo{note}{Doi:{\color{blue} \href{
  https://doi.org/10.1109/JIOT.2018.2853663}{10.1109/JIOT.2018.2853663}}}.
%Type = Article
\bibitem[{Tibshirani(1996)}]{4tibshirani1996regression}
\bibinfo{author}{Tibshirani, R.}, \bibinfo{year}{1996}.
\newblock \bibinfo{title}{Regression shrinkage and selection via the lasso}.
\newblock \bibinfo{journal}{Journal of the Royal Statistical Society: Series B
  (Methodological)} \bibinfo{volume}{58}, \bibinfo{pages}{267--288}.
\newblock \bibinfo{note}{Doi:{\color{blue}
  \href{https://doi.org/10.1111/j.2517-6161.1996.tb02080.x}{10.1111/j.2517-6161.1996.tb02080.x}}}.
%Type = Article
\bibitem[{Xu et~al.(2014)Xu, Du, Dai and Lee}]{j6xu2014regression}
\bibinfo{author}{Xu, Y.}, \bibinfo{author}{Du, J.}, \bibinfo{author}{Dai,
  L.R.}, \bibinfo{author}{Lee, C.H.}, \bibinfo{year}{2014}.
\newblock \bibinfo{title}{A regression approach to speech enhancement based on
  deep neural networks}.
\newblock \bibinfo{journal}{IEEE/ACM Transactions on Audio, Speech, and
  Language Processing} \bibinfo{volume}{23}, \bibinfo{pages}{7--19}.
\newblock \bibinfo{note}{Doi:{\color{blue}
  \href{https://doi.org/10.1109/TASLP.2014.2364452}{10.1109/TASLP.2014.2364452}}}.
%Type = Article
\bibitem[{Yagiz and Gokceoglu(2010)}]{8yagiz2010application}
\bibinfo{author}{Yagiz, S.}, \bibinfo{author}{Gokceoglu, C.},
  \bibinfo{year}{2010}.
\newblock \bibinfo{title}{Application of fuzzy inference system and nonlinear
  regression models for predicting rock brittleness}.
\newblock \bibinfo{journal}{Expert Systems with Applications}
  \bibinfo{volume}{37}, \bibinfo{pages}{2265--2272}.
\newblock \bibinfo{note}{Doi:{\color{blue}
  \href{https://doi.org/10.1016/j.eswa.2009.07.046}{10.1016/j.eswa.2009.07.046}}}.
%Type = Article
\bibitem[{Zhang et~al.(2020)Zhang, Zhao, Lin, Tan and Cheng}]{jj4zhang2020high}
\bibinfo{author}{Zhang, J.}, \bibinfo{author}{Zhao, J.}, \bibinfo{author}{Lin,
  H.}, \bibinfo{author}{Tan, Y.}, \bibinfo{author}{Cheng, J.X.},
  \bibinfo{year}{2020}.
\newblock \bibinfo{title}{High-speed chemical imaging by dense-net learning of
  femtosecond stimulated raman scattering}.
\newblock \bibinfo{journal}{The Journal of Physical Chemistry Letters}
  \bibinfo{volume}{11}, \bibinfo{pages}{8573--8578}.
\newblock \bibinfo{note}{Doi:{\color{blue} \href{
  https://doi.org/10.1021/acs.jpclett.0c01598}{10.1021/acs.jpclett.0c01598}}}.
%Type = Article
\bibitem[{Zhang et~al.(2018)Zhang, Liang, Dong, Xie and
  Cao}]{jj2zhang2018sparse}
\bibinfo{author}{Zhang, Z.}, \bibinfo{author}{Liang, X.},
  \bibinfo{author}{Dong, X.}, \bibinfo{author}{Xie, Y.}, \bibinfo{author}{Cao,
  G.}, \bibinfo{year}{2018}.
\newblock \bibinfo{title}{A sparse-view ct reconstruction method based on
  combination of densenet and deconvolution}.
\newblock \bibinfo{journal}{IEEE transactions on medical imaging}
  \bibinfo{volume}{37}, \bibinfo{pages}{1407--1417}.
\newblock \bibinfo{note}{Doi:{\color{blue} \href{
  https://doi.org/10.1109/TMI.2018.2823338}{10.1109/TMI.2018.2823338}}}.
%Type = Article
\bibitem[{Zhou et~al.(2011)Zhou, Pierre and Trudnowski}]{2zhou2011stepwise}
\bibinfo{author}{Zhou, N.}, \bibinfo{author}{Pierre, J.W.},
  \bibinfo{author}{Trudnowski, D.}, \bibinfo{year}{2011}.
\newblock \bibinfo{title}{A stepwise regression method for estimating dominant
  electromechanical modes}.
\newblock \bibinfo{journal}{IEEE Transactions on Power Systems}
  \bibinfo{volume}{27}, \bibinfo{pages}{1051--1059}.
\newblock \bibinfo{note}{Doi:{\color{blue}
  \href{https://doi.org/10.1109/TPWRS.2011.2172004}{10.1109/TPWRS.2011.2172004}}}.
%Type = Article
\bibitem[{Zou and Hastie(2005)}]{5zou2005regularization}
\bibinfo{author}{Zou, H.}, \bibinfo{author}{Hastie, T.}, \bibinfo{year}{2005}.
\newblock \bibinfo{title}{Regularization and variable selection via the elastic
  net}.
\newblock \bibinfo{journal}{Journal of the royal statistical society: series B
  (statistical methodology)} \bibinfo{volume}{67}, \bibinfo{pages}{301--320}.
\newblock \bibinfo{note}{Doi:{\color{blue}
  \href{https://doi.org/10.1111/j.1467-9868.2005.00503.x}{10.1111/j.1467-9868.2005.00503.x}}}.

\end{thebibliography}

\bio{}
\endbio

\end{document}